%% file: main.tex
\documentclass[10pt,twocolumn,letterpaper]{article}

\usepackage[pagenumbers]{iccv} 


\definecolor{iccvblue}{rgb}{0.21,0.49,0.74}
\usepackage[pagebackref,breaklinks,colorlinks,allcolors=iccvblue]{hyperref}

\def\paperID{2145} 
\def\confName{ICCV}
\def\confYear{2025}
\newcommand{\mysmall}[1]{\scriptsize{\color{gray}{#1}}}

\title{GGTalker: Talking Head Systhesis with Generalizable Gaussian Priors and Identity-Specific Adaptation}

\author{Wentao Hu$^{1*}$\quad Shunkai Li$^{2*}$\quad Ziqiao Peng$^3$\quad Haoxian Zhang$^2$ \quad Fan Shi$^2$ \\ Xiaoqiang Liu$^2$\quad Pengfei Wan$^2$\quad Di Zhang$^{2}$\quad Hui Tian$^{1 \dagger}$ \\
$^1$Beijing University of Posts and Telecommunications\\
$^2$Kuaishou Technology \quad $^3$Renmin University of China
}
\begin{document}

\twocolumn[{
\maketitle
\begin{center}
    \captionsetup{type=figure}
    \vspace{-1.7em}
    \includegraphics[width=0.96\textwidth]{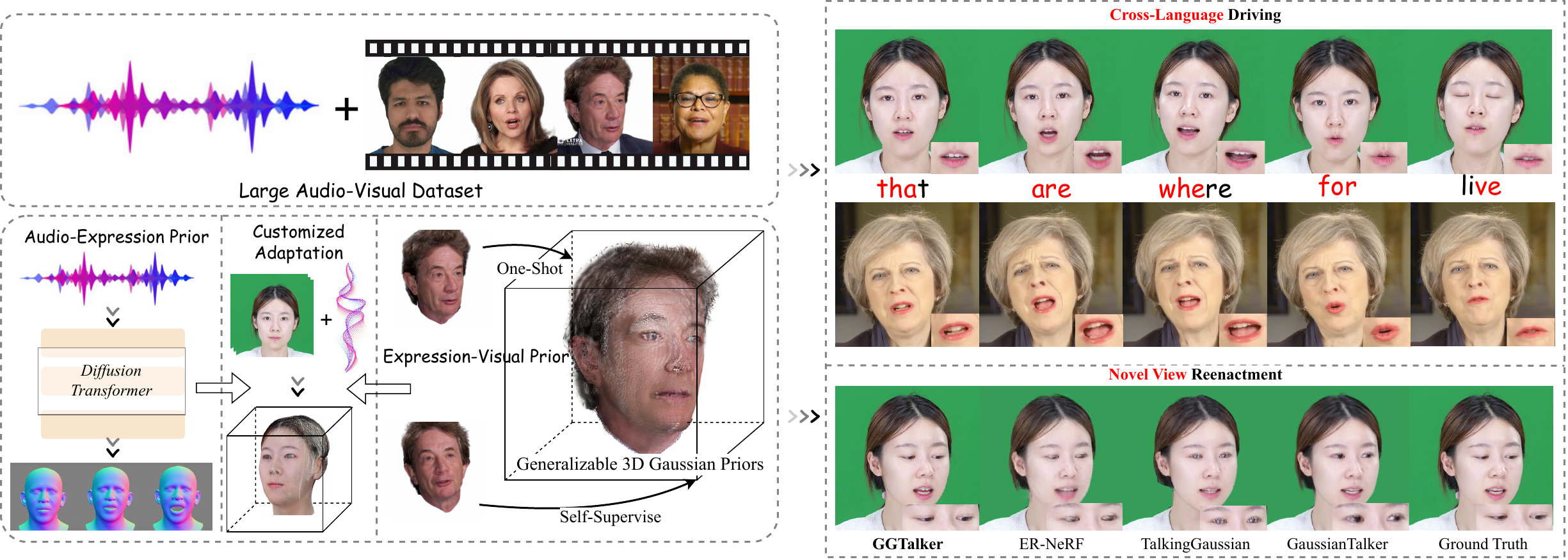}
    \vspace{-0.5em}
    \captionof{figure}{The proposed GGTalker introduces generalizable Gaussian head priors from large-scale datasets to identity-specific talking head. With the Prior-Adaption approach, we implement novel angle reenactment and cross-language driving to generate high-fidelity results.}
    \label{fig:teaser}

\end{center}
}]

\renewcommand{\thefootnote}{\fnsymbol{footnote}} 
\footnotetext[2]{Corresponding author.}
\footnotetext[1]{Equal contribution.}

\input{sec/0_abstract}    
\input{sec/1_intro}
\input{sec/2_relatedwork}
\input{sec/3_method}

\input{sec/4_experiments}
\input{sec/5_conclusion}
\input{sec/6_Acknowledgement}

{
    \small
    \bibliographystyle{ieeenat_fullname}
    \bibliography{main}
}

\end{document}

%% file: sec/0_abstract.tex
\begin{abstract}
Creating high-quality, generalizable speech-driven 3D talking heads remains a persistent challenge. Previous methods achieve satisfactory results for fixed viewpoints and small-scale audio variations, but they struggle with large head rotations and out-of-distribution (OOD) audio. Moreover, they are constrained by the need for time-consuming, identity-specific training. We believe the core issue lies in the lack of sufficient 3D priors, which limits the extrapolation capabilities of synthesized talking heads. To address this, we propose GGTalker, which synthesizes talking heads through a combination of generalizable priors and identity-specific adaptation. We introduce a two-stage Prior-Adaptation training strategy to learn Gaussian head priors and adapt to individual characteristics. We train Audio-Expression and Expression-Visual priors to capture the universal patterns of lip movements and the general distribution of head textures. During the Customized Adaptation, individual speaking styles and texture details are precisely modeled. Additionally, we introduce a color MLP to generate fine-grained, motion-aligned textures and a Body Inpainter to blend rendered results with the background, producing indistinguishable, photorealistic video frames. Comprehensive experiments show that GGTalker achieves state-of-the-art performance in rendering quality, 3D consistency, lip-sync accuracy, and training efficiency.
\end{abstract}

%% file: sec/1_intro.tex
\section{Introduction}
\label{sec:intro}
With the increasing demand for virtual reality~\cite{morishima1998real}, digital agents~\cite{thies2020neural} and other applications, creating realistic, vivid and expressive talking heads has become a crucial task of computer vision. Significant progress has been made in this area, with numerous remarkable achievements. 
In particular, with the development of efficient 3D rendering technologies such as Neural Radiance Fields (NeRF)~\cite{mildenhall2021nerf} and 3D Gaussian Splatting (3DGS)~\cite{kerbl20233d} in recent years, 3D-based talking heads~\cite{guo2021ad, li2023efficient, peng2024synctalk, ye2023geneface, li2024talkinggaussian, cho2024gaussiantalker, aneja2024gaussianspeech, yu2024gaussiantalker, peng2025synctalk++, li2025instag, li2024er, ye2024real3d, ye2024mimictalk} are gradually gaining more attention than 2D-based ones~\cite{prajwal2020lip, zhang2023dinet, sinha2020identity, cheng2022videoretalking, zhang2023sadtalker, shen2023difftalk}, thanks to their strong identity consistency and fast rendering speed.

However, current 3D-based talking head methods generally lack generalizability, as reflected in the following aspects:
(1) They only support inference on audio similar to the training data, performing poorly on out-of-distribution (OOD) audio.
(2) Current methods struggle to synthesize large head rotations, such as side profiles, upward or downward head movements, limiting the expressiveness of talking head synthesis. We believe this limitation stems from most existing methods training from scratch for each specific identity~\cite{guo2021ad, li2023efficient, peng2024synctalk, li2024talkinggaussian, cho2024gaussiantalker, aneja2024gaussianspeech}, without incorporating 3D head priors. As a result, the generated talking heads overfit the training data and lack extrapolation ability. Moreover, training from scratch for each identity is inefficient, often requiring several hours per session~\cite{li2023efficient, peng2024synctalk, ye2023geneface, li2024talkinggaussian, cho2024gaussiantalker, aneja2024gaussianspeech, yu2024gaussiantalker}. Some methods even demand costly multi-view synchronized videos~\cite{aneja2024gaussianspeech, he2024emotalk3d}. The inefficiency further hinders the widespread adoption of talking head technology.

To address this, we propose \textbf{G}eneralizable \textbf{G}aussian \textbf{Talker} (\textbf{GGTalker}), which introduces facial priors from large-scale datasets into identity-specific adaptation. We argue that the shape and texture of all human heads, as well as the correlation between audio and lip movements, follow general patterns that can be learned from large-scale data. For a specific identity, these patterns can be fine-tuned to adapt individual speaking habits and facial details. We believe this Prior-Adaptation training approach mitigates overfitting to the training set, enabling adaptation to large head rotations and OOD audio. At the same time, leveraging well-trained priors makes this approach significantly more efficient than training from scratch.

Specifically, 
to introduce facial shape priors and explicitly control head poses, we use FLAME~\cite{li2017learning} as the intermediate representation for the talking head~\cite{peng2025dualtalk} and rig the Gaussians to the FLAME mesh~\cite{qian2024gaussianavatars}. 
To learn Audio-Expression Priors, we train a diffusion transformer model on large-scale datasets to predict audio-conditioned expressions. For Expression-Visual Priors, we leverage the UV layout of the canonical mesh to elegantly predict 3D priors in the UV space. We use an Identity-Gaussian Generator to predict reasonable initial 3D Gaussian head from a single reference image and introduce a source-target self-supervised training paradigm to accommodate cross-identity monocular and multi-view datasets. To adapt to the target individual’s speaking style and texture details, we jointly fine-tune the lip motion prior and texture prior on specific data. Simultaneously, we optimize the FLAME parameters to mitigate the errors caused by the current imperfect monocular tracking algorithms. We further introduce a color MLP to generate motion-aligned sharp textures and a Body Inpainter to blend rendered results with the torso and background, producing indistinguishable, photorealistic video results. Comprehensive experiments show that GGTalker outperforms in rendering quality, 3D consistency, lip-sync accuracy, training efficiency and inference speed.

Our contributions can be summarized as follows:
\begin{itemize}
\item[$\bullet$] We propose GGTalker, a two-stage Prior-Adaptation method focused on synthesizing high-fidelity and generalizable 3D talking heads.

\item[$\bullet$] We use FLAME to explicitly control facial motions. We train Audio-Expression and Expression-Visual priors on large datasets to learn the general patterns of audio-driven lip movements and head textures.

\item[$\bullet$] We adapt well-trained priors to target identity’s speaking style and texture details. We further introduce a color MLP and a Body Inpainter to generate motion-aligned fine textures and lifelike results.
\end{itemize}

%% file: sec/2_relatedwork.tex
\begin{figure*}
\vspace{-1em}
\begin{center}
   \includegraphics[width=1.\linewidth]{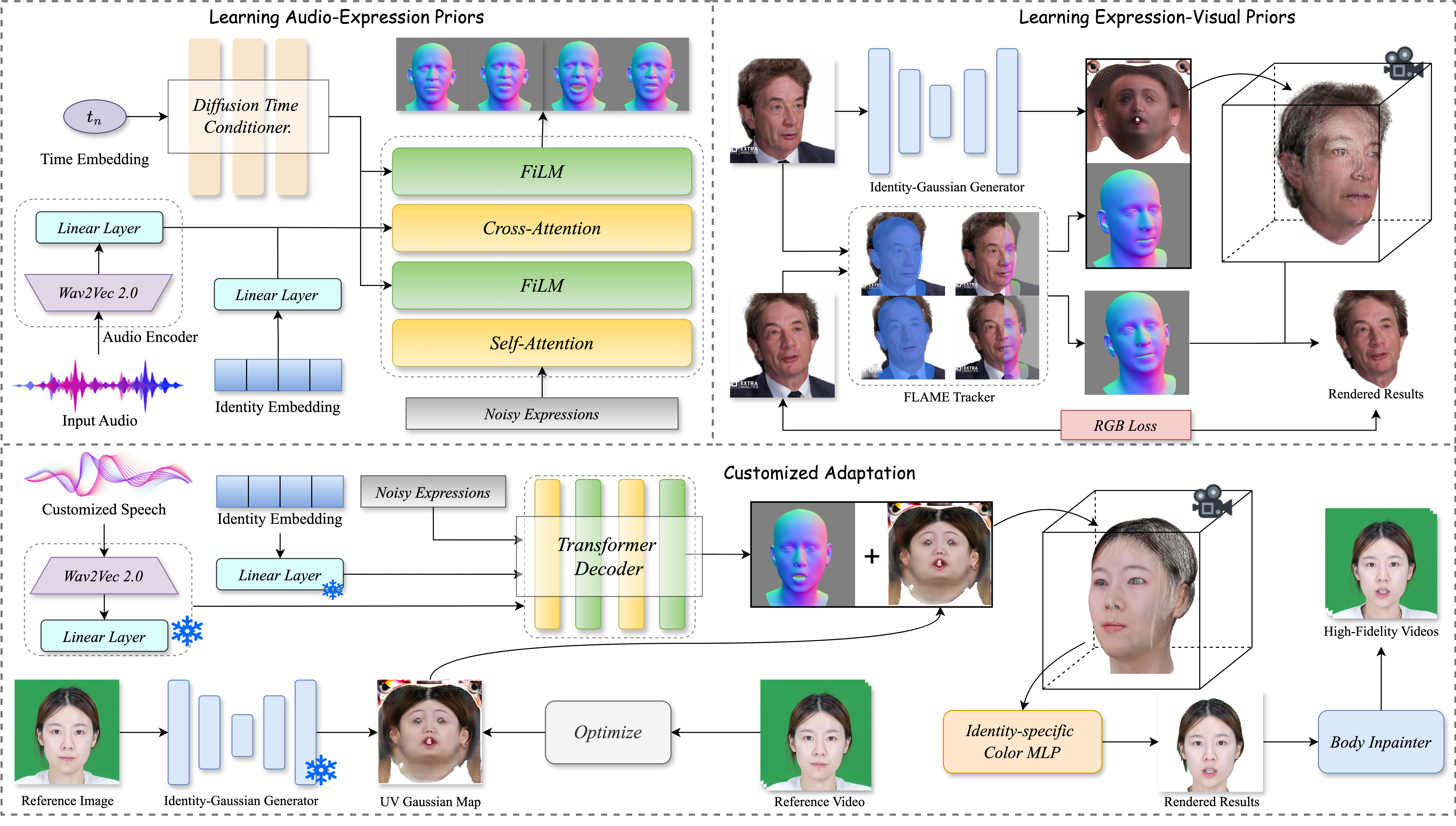}
\end{center}
\vspace{-1em}
   \caption{\textbf{Overview of GGTalker.} We introduce a two-stage Prior-Adaptation approach for talking head synthesis. We use a conditional diffusion transformer to predict plausible facial movements from an audio sequence. For texture generation, we leverage the UV layout of the canonical mesh and use an Identity-Gaussian Generator to predict a UV Gaussian map from a single reference image, initializing a reasonable Gaussian distribution for the head. We adapt well-trained priors to target individual’s speaking style and texture details. We further introduce a color MLP and a Body Inpainter to generate motion-aligned fine textures and lifelike results.}
\label{fig:pipeline}
\vspace{-0.5em}
\end{figure*}

\section{Related Work}
\subsection{2D-Based Talking Head}
2D-based methods, including  Generative Adversarial Networks (GAN) and diffusion models, are commonly used for synthesizing talking heads due to their powerful generative capabilities~\cite{zhou2021pose,song2022everybody, meshry2021learned, peng2025omnisync}. For instance, Wav2Lip~\cite{prajwal2020lip} introduces a lip synchronization expert to supervise lip movements, enforcing the consistency of lip movements with the audio. IP-LAP~\cite{zhong2023identity} proposes a two-stage framework consisting of audio-to-landmark generation and landmark-to-video rendering procedures. However, these methods generate only the mouth region or the low half of face, which can lead to uncoordinated facial movements and artifacts. Apart from video-based methods, some approaches enable speech-driven animation from a single image~\cite{zhang2023sadtalker, xu2024hallo, chen2024echomimic}. For example, Hallo~\cite{xu2024hallo} introduces a hierarchical cross-attention mechanism to augment the correlation between audio inputs and non-identity-related motions. However, relying solely on a single reference image making it difficult to maintain the identity consistently and talking style.

In contrast, GGTalker uses 3D Gaussians to explicitly represent talking heads, leading to exceptional performance in maintaining subject identity consistency and detail preservation. Simultaneously, its training and rendering speed is significantly superior to 2D-based methods. 

\subsection{3D-Based Talking Head}

With the rise of 3D rendering methods like NeRF~\cite{mildenhall2021nerf} and 3DGS~\cite{kerbl20233d}, their high rendering speed and quality have garnered widespread attention in the talking head domain. 
Earlier works have incorporated NeRF~\cite{guo2021ad, li2024er, ye2023geneface, peng2024synctalk} into talking head tasks. For example, RAD-NeRF~\cite{tang2022real} optimized inference for real-time performance, while ER-NeRF~\cite{li2024er} introduced a three-plane hash encoder to accelerate rendering. SyncTalk~\cite{peng2024synctalk} introduced synchronization modules to improve alignment between audio and generated facial expressions.
Recent research has shifted towards Gaussian Splatting~\cite{yu2024gaussiantalker,cho2024gaussiantalker,li2024talkinggaussian} due to its rendering efficiency and multi-view consistency. GaussianTalker~\cite{cho2024gaussiantalker} uses point-based Gaussian Splatting, encoding 3D Gaussian attributes into a shared implicit representation, which is merged with audio features to manipulate each Gaussian attribute. 

However, these methods are typically trained on 3-5 minutes of video data for specific individuals without incorporating additional priors. As a result, generating large-scale head movements often leads to artifacts and distortions due to insufficient 3D information, and lip movements may not sync for certain characters. To address these issues, we introduce two additional priors—Audio-Expression Priors and Expression-Visual Priors—to improve the quality and fidelity of generated talking heads.



%% file: sec/3_method.tex
\section{Method}
As shown in Fig.~\ref{fig:pipeline}, GGTalker consists of 3 parts: a) Audio-Expression model for generating expression sequences from audio features, b) Expression-Visual model for predicting coarse head textures, and c) Customized Adaptation for identity-specific facial textures and speaking habits, producing photorealistic results.

\subsection{Preliminary}
\label{sec:3.1}
3DGS represents 3D scene with a set of Gaussians~\cite{kerbl20233d}. Each Gaussian primitive $\mathcal{G}^i$ is described by a global space position $\mu \in \mathbb{R}^3$, scale $\mathbf{s} \in \mathbb{R}^3$, rotation (parameterized as quaternion) $\mathbf{r} \in \mathbb{R}^4$, color $\mathbf{SH} \in \mathbb{R}^3$, and opacity $\sigma \in \mathbb{R}$.
\begin{align}
    \mathcal{G}^i = \{\mathbf{\mu}, \mathbf{s}, \mathbf{r}, \mathbf{SH}, \sigma\} \label{eq:global}
\end{align}

Given the differentiable rasterizer $\mathcal{R}$ and camera view $\pi$, 3D Gaussians $\mathcal{G}$ can be efficiently rendered to an image $I$ :
\begin{align}
    I = \mathcal{R}(\mathcal{G}, \pi)  \label{eq:rendering}
\end{align}

\subsection{Audio-Expression Priors}



\noindent\textbf{Audio Condition Encoder.}  
We employ a state-of-the-art pre-trained speech model Wav2Vec 2.0~\cite{baevski2020wav2vec} to encode the audio signal. Given the input audio sequence $\mathbf{A} = [\mathbf{a}_1, \dots, \mathbf{a}_T]$, the audio condition encoder processes it by first projecting each feature $\mathbf{a}_t \in \mathbb{R}^{1280}$ into a lower-dimensional space ($\mathbb{R}^{d}$, where $d= 512$) via a learned linear layer. A shallow Transformer encoder refines temporal dependencies to better capture expressive variations. Additionally, we introduce a identity embedding $\mathbf{I} \in \mathbb{R}^{64}$ to retrieve the the sample’s speaker ID, which is transformed via linear layers and concatenated with the projected audio features. 
The final conditioning sequence consists of per-frame tokens $\mathbf{C} \in \mathbb{R}^{T \times d}$ and a global conditioning vector $\mathbf{h} \in \mathbb{R}^{d}$. After layer normalization and encoding, we obtain the enriched condition sequence $\mathbf{C}^{\prime} = [\mathbf{c}^{\prime}_1, …,\mathbf{c}^{\prime}_T]$ and an aggregated conditioning vector:
\begin{equation}
\bar{\mathbf{c}} = \frac{1}{T} \sum_{t=1}^{T} \mathbf{c}^{\prime}_t.
\end{equation}
The enriched condition sequence $\mathbf{C}^{\prime}$ is used for frame-level conditioning, while $\bar{\mathbf{c}}$ provides a global context to guide expression generation.

\noindent\textbf{Diffusion Time Conditioner.}
To model the sequential nature of facial expressions, we employ a denoising diffusion probabilistic model (DDPM) to iteratively refine expression sequences. During training, each sample is assigned a random diffusion timestep $n$, where $n=N$ corresponds to nearly pure noise sampled from a Gaussian prior, and $n=0$ represents the original clean expression sequence. The model learns to predict the denoised expression $\mathbf{z}_t$ from its noisy counterpart. The timestep $n$ is encoded using sinusoidal position embeddings and transformed via an MLP into a time embedding $\mathbf{t}_n \in \mathbb{R}^{d}$. This embedding is incorporated into the model through two mechanisms: (1) a FiLM modulation where $\mathbf{t}_n$ is linearly combined with $\bar{\mathbf{c}}$, and (2) a tokenized representation where $\mathbf{t}_n$ is mapped into additional time tokens appended to $\mathbf{C}^{\prime}$. These conditioning mechanisms provide a continuous signal to inform the model of the noise level at each step, allowing it to adaptively refine the expression sequence.

\noindent\textbf{Transformer Decoder}
The Transformer decoder is responsible for denoising and generating expressive facial motion sequences. Given a noisy input sequence $\mathbf{Z} = [\mathbf{z}_1, …, \mathbf{z}_T]$, where each $\mathbf{z}_t \in \mathbb{R}^{K}$ represents a noisy expression parameter vector at time $t$, we first project the inputs to the latent space and apply learned positional encodings. The decoder consists of $L=8$ stacked Transformer layers, each incorporating self-attention and cross-attention mechanisms. Self-attention captures temporal dependencies across frames, ensuring smooth and coherent expressions, while cross-attention aligns the generated expressions with the conditioning tokens $\mathbf{C}^{\prime}$ to enforce synchronization with the input speech features. FiLM conditioning is applied at each decoder layer to modulate activations using the global condition $\bar{\mathbf{c}}$ and the diffusion timestep embedding $\mathbf{t}_n$. This design enables the model to balance local frame-wise synchronization with global temporal consistency.

To enhance generalization, we apply classifier-free guidance by randomly setting $\mathbf{C}^{\prime}$ to zero with probability $p=0.1$ during training. This encourages the model to learn an implicit prior over plausible facial expressions, making it more robust to variations in audio input. The decoder’s final outputs are mapped to the predicted expression sequence $\hat{\mathcal{F}}_{exp} = [\hat{\mathbf{e}}_1, \hat{\mathbf{e}}_2, \dots, \hat{\mathbf{e}}_T]$, which can be applied to a FLAME-based 3D face model for animation. The iterative diffusion process ensures diverse and high-fidelity expression synthesis. Empirically, we find that predicting residual noise or directly estimating the denoised expression at each step leads to improved synchronization between audio and facial expressions:
\begin{equation}
\hat{\mathbf{e}}_t = f_{\theta} (\mathbf{z}_t, \mathbf{C}^{\prime}, \bar{\mathbf{c}}, \mathbf{t}_n).
\end{equation}
This formulation enhances robustness and improves the model's ability to generate accurate and expressive facial motions.

\noindent\textbf{Loss Functions.} 
To obtain the precise expressions, we conduct the L2 regularization $\mathcal{L}_{exp}$. To mitigate the flicker between adjacent frames, We use a temporal constraint $\mathcal{L}_{temp}$ and optimize $n=10$ frames within a sliding window:
\begin{equation}
\begin{aligned}
 \mathcal{L}_{temp} = \sum_{i=1}^{|n|}  \big\| \boldsymbol{\mathcal{F}}_{exp}^{i+1} - \boldsymbol{\mathcal{F}}_{exp}^{i}  \big\| _{\epsilon},\\
 \mathcal{L}_{exp} = \sum_{i=1}^{|n|}  \big\|  \boldsymbol{\mathcal{F}}_{exp}^{i} - \boldsymbol{\mathcal{F}}_{exp} \big\|_2,
\end{aligned}
\end{equation}
where $\|.\|_2$ and $\|.\|_{\epsilon}$ refers to L2 and Huber loss. The total loss is defined as: 
\begin{equation}
    \mathcal{L}_{\text{A2E}} = \lambda_{temp} \mathcal{L}_{temp} + \lambda_{exp} \mathcal{L}_{exp}. 
\end{equation}

\subsection{Expression-Visual Priors}

\noindent\textbf{Gaussian Rigging.} 
As a point-based representation, 3DGS is inherently unstructured, making it difficult to control. 
Thus, we rig the Gaussians to the FLAME mesh to ensure a clear and consistent 3D structure~\cite{qian2024gaussianavatars}.
Specifically, we define the center $\boldsymbol{C}^i$ of each triangle as the origin of its local coordinate system.  The triangle’s scale is determined by computing the average length $\boldsymbol{l}^i$ of an edge and its perpendicular, along with the negative correlation value $\boldsymbol{k}^i$ of $\boldsymbol{l}^i$.  The triangle’s orientation in world space is represented by a rotation matrix $\boldsymbol{R}^i$, obtained through the cross product of the edge direction and normal vectors.  During rendering, the global Gaussian attributes $\mathcal{G}^i$ are calculated by local attributes $\mathcal{G}^i_l$ and transformation $\mathcal{W}$:
\begin{equation}
\mathcal{G}^i=\left\{
\begin{aligned}
\mathbf{\mu} &= \boldsymbol{k}^i \boldsymbol{R}^i \mathbf{\mu}_{l} + \boldsymbol{C}^i \\
\mathbf{s} &= \boldsymbol{k}^i \mathbf{s}_{l} \\
\mathbf{r} &= \boldsymbol{R}^i \mathbf{r}_{l} \\
\mathbf{SH} &= \mathbf{SH}_{l} \\
\mathbf{\sigma} &= \mathbf{\sigma}_{l}
\end{aligned}
\right. \label{eq:local2global}
\end{equation}

\noindent\textbf{Self-Supervised Learning.} 
Despite using the FLAME mesh to constrain the spatial distribution of Gaussians, we found that simply fitting a identity-specific training video does not yield satisfactory results. The generated talking heads appear flawless when facing forward, but artifacts and holes emerge during head-turning motions or when viewed from novel angles, as shown in Fig.~\ref{fig:ablation}. This is because monocular training videos lack sufficient 3D facial information, leaving the Gaussians on the sides of the head under-constrained. 
However, we argue that all human heads share a similar texture pattern, and a network should be able to predict reasonable Gaussian patterns after training on large amounts of data. Inspired by~\cite{zheng2024headgap, chu2024generalizable}, we learn 3D Gaussian Priors from large datasets using self-supervised learning. 

\begin{figure}
\begin{center}
   \includegraphics[width=1.\linewidth]{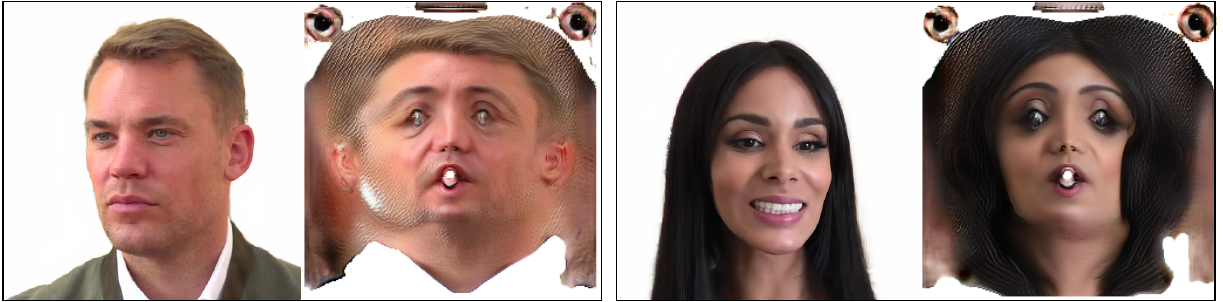}
\end{center}
\vspace{-1em}
   \caption{\textbf{Visualization of the Gaussian color attributes} on UV map $M$. We leverage the UV layout of the canonical mesh to map 2D identity features into 3D space.}
\label{fig:UV_color}
\vspace{-1em}
\end{figure}

Specifically, given a collection of videos for a specific identity, we randomly select two frames as the source and target image $<I_{src}, I_{tgt}>$. We assume that a clear, front-facing image with distinct facial features can capture most of the identity information. Therefore, we use an Identity-Gaussian Generator $G$ to predict a UV Gaussian map $M \in \mathbb{R}^{H\times W\times 14}$ from $I_{src}$. Each pixel in $M$ is treated as a Gaussian, where the depth dimension corresponds to the intrinsic 14-dimensional parameters of 3DGS. We visualize the 3-dimensional of $M$, which represent the color of Gaussians, as shown in Fig.~\ref{fig:UV_color}. To obtain the actual 3D Gaussian primitives for rasterization, we simply sample the UV map $M$ uniformly and place the primitives relative to the canonical mesh~\cite{kirschstein2024gghead}:
\begin{equation}
    \mathcal{G} =  \mathcal{W}(\mathcal{G}_l) = \textsc{Sample}(M) = \textsc{Sample}(G(I_{src}))
\end{equation}
By leveraging the UV layout of the canonical mesh, we elegantly achieve a forward conversion from a 2D image to a 3D Gaussian head — a process that typically requires per-frame fitting in previous methods~\cite{li2024talkinggaussian, yu2024gaussiantalker, cho2024gaussiantalker}. Additionally, compared to initializing Gaussians directly on mesh vertices or faces~\cite{qian2024gaussianavatars}, our method results in a more uniform Gaussian distribution~\cite{xiang2024flashavatar}. To supervise the predicted 3D Gaussian head $\mathcal{G}$, we use 
the target image $I_{tgt}$ to drive $\mathcal{G}$, rendering it from the camera view $\pi_{tgt}$ of $I_{tgt}$:
\begin{equation}
    \hat{I}_{tgt} = \mathcal{R}(\mathcal{B}(\mathcal{W}(\mathcal{G}_l), \mathcal{F}_{exp}, \mathcal{F}_{pose}), \pi_{tgt}), 
\end{equation}
where $\mathcal{F}_{exp}$ and $ \mathcal{F}_{pose}$ denote the pose and expression parameters of $I_{tgt}$. $\mathcal{B}$ denotes the FLAME blendshapes. The rendered result $\hat{I}_{tgt}$ is then supervised by ground truth $I_{tgt}$. In this way, we achieve self-supervised 3D prior learning.

\noindent\textbf{Loss Functions.} 
First, since we rig Gaussians to the FLAME mesh, we prevent them from drifting too far from their corresponding triangles. Therefore, we introduce position regularization:

\begin{equation}
\mathcal{L}_{\mu} = \Vert \mu_{l} \Vert_2
\end{equation}

As for RGB-level supervision, we use a combination of L1 loss, SSIM loss, and a VGG-based perceptual loss. The total loss function is defined as:

\begin{equation}
\mathcal{L}_{\text{E2V}} = \lambda_{\text{L1}} \mathcal{L}_{\text{L1}} + \lambda_{\text{SSIM}} \mathcal{L}_{\text{SSIM}} + \lambda_{\text{vgg}} \mathcal{L}_{\text{vgg}} + \lambda_{\mu} \mathcal{L}_{\mu}
\end{equation}

\begin{table*}[]
\setlength\tabcolsep{5pt}
\begin{center}
\resizebox{\linewidth}{!}{
\begin{tabular}{lcccc|ccc|cc} 
\toprule
Methods               & PSNR ↑           & LPIPS ↓         & SSIM ↑& FID ↓           & LMD ↓& AUE ↓& LSE-C ↑& Training ↓& FPS ↑\\ 
\midrule
Wav2Lip \mysmall{(MM 20 \cite{prajwal2020lip})}& 31.971& 0.0767& 0.9512& 16.891& 5.248& 3.924& \textbf{8.873} \footnotemark{}& -& 19.6\\
DINet \mysmall{(AAAI 23 \cite{zhang2023dinet})}& 28.348& 0.0698& 0.9250& 10.714& 5.836& 4.847& 5.890& -& 23.7\\
IP-LAP \mysmall{(CVPR 23 \cite{zhong2023identity})}& 32.784& 0.0488& \underline{0.9785}& 8.860& 3.931& 3.184& 4.067& -& 3.27\\ 
\midrule
AD-NeRF \mysmall{(ICCV 21 \cite{guo2021ad})}& 26.183& 0.1512& 0.9137& 29.398& 4.134& 6.238& 3.947& 30h& 0.15\\
 RAD-NeRF \mysmall{(arXiv 22 \cite{tang2022real})}& 29.486& 0.0572& 0.9329& 9.634& 4.086& 5.247& 5.133& 4h&36.6\\
 ER-NeRF \mysmall{(ICCV 23 \cite{li2023efficient})}& 30.438& 0.0408& 0.9331& 5.516& 4.014& 4.774& 5.008& 5h&38.2\\
GeneFace \mysmall{(ICLR 23 \cite{ye2023geneface})}& 26.756& 0.1197& 0.8792& 21.086& 5.139& 5.826& 5.117& 13h& 0.18\\
 GeneFace++ \mysmall{(arxiv 23 \cite{ye2023geneface++})}& 27.490& 0.1056& 0.8971& 19.435& 5.383& 5.353& 5.746& 5h&42.5\\
 SyncTalk \mysmall{(CVPR 24 \cite{peng2024synctalk})}& 32.545& \underline{0.0334}& 0.9630& 6.820& \underline{2.963}& 3.618& 7.693& 5h&31.9\\ 
 \midrule
TalkingGaussian \mysmall{(ECCV 24 \cite{li2024talkinggaussian})}&31.714&0.0396&\underline{0.9664}&6.577&3.460&\underline{2.793}&6.082&\underline{0.5h}&\underline{76.3}\\
GaussianTalker \mysmall{(MM 24 \cite{cho2024gaussiantalker})}&\underline{32.941}&0.0531&0.9531&\underline{6.392}&3.061&2.980&6.109&1h&72.8\\
\midrule
Ours& \textbf{35.203}&\textbf{0.0281}&\textbf{0.9816}&\textbf{4.624}&\textbf{2.328}&\textbf{2.171}&\underline{8.210}&\textbf{0.3h}&\textbf{120}\\ 
\bottomrule
\end{tabular}}
\end{center}
\vspace{-1em}
\caption{\textbf{Quantitative results of self-reenactment.} We achieve state-of-the-art performance on most metrics. We highlight \textbf{best} and \underline{second-best} results.}
\label{tab:main_compare}
\vspace{-1em}
\end{table*}

\subsection{Customized Adaptation} 

\noindent\textbf{Expression-Visual Fine-tuning} During the Expression-Visual Prior stage, our Identity-Gaussian Generator predicts a reasonable texture prior from a single image. However, in essence, a single image cannot capture complete and precise identity information. Relying solely on textures provided by the Expression-Visual Prior yields rough and smooth results (shown in Fig.~\ref{fig:ablation}). Therefore, we introduce an adaptation method. First, we select a front-facing reference image $I_{ref}$ with clear facial features based on facial landmarks — a step already accomplished during FLAME tracking. The reference image $I_{ref}$ is fed into the trained Identity-Gaussian Generator to produce a coarse UV Gaussian map $\hat{M}_{\text{id}}$:
\begin{equation}
    \hat{M}_{\text{id}} = G(I_{ref})
\end{equation}
We then combine $\hat{M}_{\text{id}}$ with the tracked FLAME parameters of the full training video, and the rasterized rendering results are supervised by ground truth:
\begin{equation}
    \hat{I}_{train} = \mathcal{R}(\mathcal{B}(\mathcal{W}(\textsc{Sample}(\hat{M}_{\text{id}})), \mathcal{F}_{exp}, \mathcal{F}_{pose}), \pi) 
\end{equation}
Note that we initially freeze the FLAME parameters and optimize only $\hat{M}_{\text{id}}$. Later, we jointly optimize both $\hat{M}_{\text{id}}$ and FLAME parameters. This approach is motivated by the limitations of current monocular tracking algorithms, which primarily minimize the error between CG rendering results and ground truth frames~\cite{qian2024versatile, feng2021learning, danvevcek2022emoca}. However, mesh shading inherently introduces inevitable errors. We argue that our mesh-rigged Gaussians can be viewed as a more advanced form of mesh shading. Therefore, we jointly fine-tune $\hat{M}_{\text{id}}$ and FLAME parameters to optimize the tracking results of existing algorithms.

\noindent\textbf{Audio-Expression Fine-tuning} 
To adapt to the speaking style of a specific identity, such as the amplitude of mouth opening, we fine-tune the Audio-Expression model on their specific data. Fine-tuning starts from the multi-speaker pre-trained weights, continuing the optimization of model parameters $\theta$ using the same total loss function.
Since personal datasets are typically limited, we employ a low learning rate 
and apply early stopping based on validation loss to prevent overfitting. To maintain the generalizability of audio features, we freeze the audio encoder $f_{\text{audio}}(\mathbf{A})$, ensuring that the extracted features $\mathbf{C}^{\prime}$ remain stable:


\begin{equation}
\mathbf{C}^{\prime} = f_{\text{cond}}(\mathbf{A}, \mathbf{I}),
\end{equation}

During fine-tuning, we update the conditional encoder $f_{\text{cond}}$ and Transformer decoder $f_{\theta}$ while keeping the audio encoder fixed:

\begin{equation}
\hat{\mathbf{e}}_t = f_{\theta} (\mathbf{z}_t, \mathbf{C}^{\prime}, \bar{\mathbf{c}}, \mathbf{t}_n).
\end{equation}

Additionally, the target identity’s embedding $\mathbf{I}$ is optimized jointly, refining its representation to better capture identity-specific facial expressions. This adaptation enables the model to better align with individual idiosyncrasies while preserving the overall consistency of speech-driven facial animation.

\noindent\textbf{Color Fine-tuning}
During the Expression-Visual Prior and UV map fine-tuning stage, Gaussian attributes remians static in local space. However, even for the same identity, textures are not static across different motions. For example, certain expressions may cause the brows to furrow, while others may lift the eyebrows. Fixing the texture during the Customized Adaptation stage would prevent the representation of dynamic micro-expressions. To address this, we use a lightweight MLP $\mathcal{M}_{\text{SH}}$ to fine-tune the color attributes of Gaussians based on pose and expression, allowing us to render sharper textures: 
\begin{equation}
    \mathbf{SH}_{l} = \mathcal{M}_{\text{SH}} ( \hat{\mathbf{SH}_{l}}, \mathcal{F}_{exp}, \mathcal{F}_{pose})
\end{equation}

\noindent\textbf{Body Inpainter. }
Since FLAME only models the head region, we limit our synthesis to the head area. To achieve better visual quality and practical applicability, we use a Body Inpainter $\mathcal{I}$ to blend the torso and background back into the rendered results. The Body Inpainter is a lightweight U-Net that helps avoid the artifacts that commonly arise from hard blending at the intersection regions. It takes as input the rendered result $I_{res}$ and the background image with a slightly dilated facial region removed, outputting highly realistic video frames $I_{vid}$ that are nearly indistinguishable from real footage.
\begin{equation}
    I_{vid} = \mathcal{I} (I_{res}, (1-\textsc{Dilate}(\mathbf{M}))I_{ori}),
\end{equation}
where $\mathbf{M}$ denotes the face mask, and $I_{ori}$ denotes original frame. Our Body Inpainter is pre-trained alongside the Expression-Visual stage and jointly fine-tuned with $\mathcal{M}_{\text{expr}}$, $\mathcal{F}_{exp}$, $\mathcal{F}_{exp}$, $\hat{M}_{\text{id}}$, $\mathcal{M}_{\text{SH}}$ in the final stage.

\noindent\textbf{Loss Functions.} 
Except for individually Audio-Expression fine-tuning, we supervise the rendered results at RGB-level, using the same losses as the Expression-Visual Prior stage: 
\begin{equation}
\mathcal{L} = \lambda_{\text{L1}} \mathcal{L}_{\text{L1}} + \lambda_{\text{SSIM}} \mathcal{L}_{\text{SSIM}} + \lambda_{\text{vgg}} \mathcal{L}_{\text{vgg}}
\end{equation}

%% file: sec/4_experiments.tex
\begin{table}[]
\resizebox{\linewidth}{!}{
\begin{tabular}{lcccc}
\toprule
 & \multicolumn{2}{c}{Cross-Identity}       & \multicolumn{2}{c}{Cross-Language}       \\ \cmidrule(l){2-5} 
Method                        & LSE-D ↓         & LSE-C ↑         & LSE-D ↓         & LSE-C ↑         \\ \midrule

IP-LAP \mysmall{(CVPR 23 \cite{zhong2023identity})}                                        &  \underline{8.583}& 5.389&  \underline{9.569}& 5.041\\
ER-NeRF \mysmall{(ICCV 23 \cite{li2023efficient})}                                     & 10.273&  3.216& 10.825&  3.051\\
GeneFace \mysmall{(ICLR 23 \cite{ye2023geneface})}                                      & 9.606& 4.014& 10.072& 3.832\\
SyncTalk \mysmall{(CVPR 23 \cite{peng2024synctalk})}                                     & 8.732& \underline{5.640}& 9.756& \underline{5.301}\\
TalkingGaussian \mysmall{(ECCV 24 \cite{li2024talkinggaussian})}                                      & 9.501& 4.344& 9.831& 3.118\\
GaussianTalker \mysmall{(MM 24 \cite{cho2024gaussiantalker})}                                      & 10.105& 4.038& 10.806& 3.499\\
\midrule
Ours                                     & \textbf{8.051}& \textbf{6.268}& \textbf{8.923}& \textbf{5.769}\\ \bottomrule
\end{tabular}}
\vspace{-1em}
\caption{\textbf{The quantitative results of the lip synchronization on OOD audio.} We separately evaluate the cross-identity and cross-language settings, then highlight \textbf{best} and \underline{second-best} results.}
\label{tab:OOD_compare}
\vspace{-1em}
\end{table}

\begin{figure*}
\vspace{-1em}
\begin{center}
   \includegraphics[width=1.\linewidth]{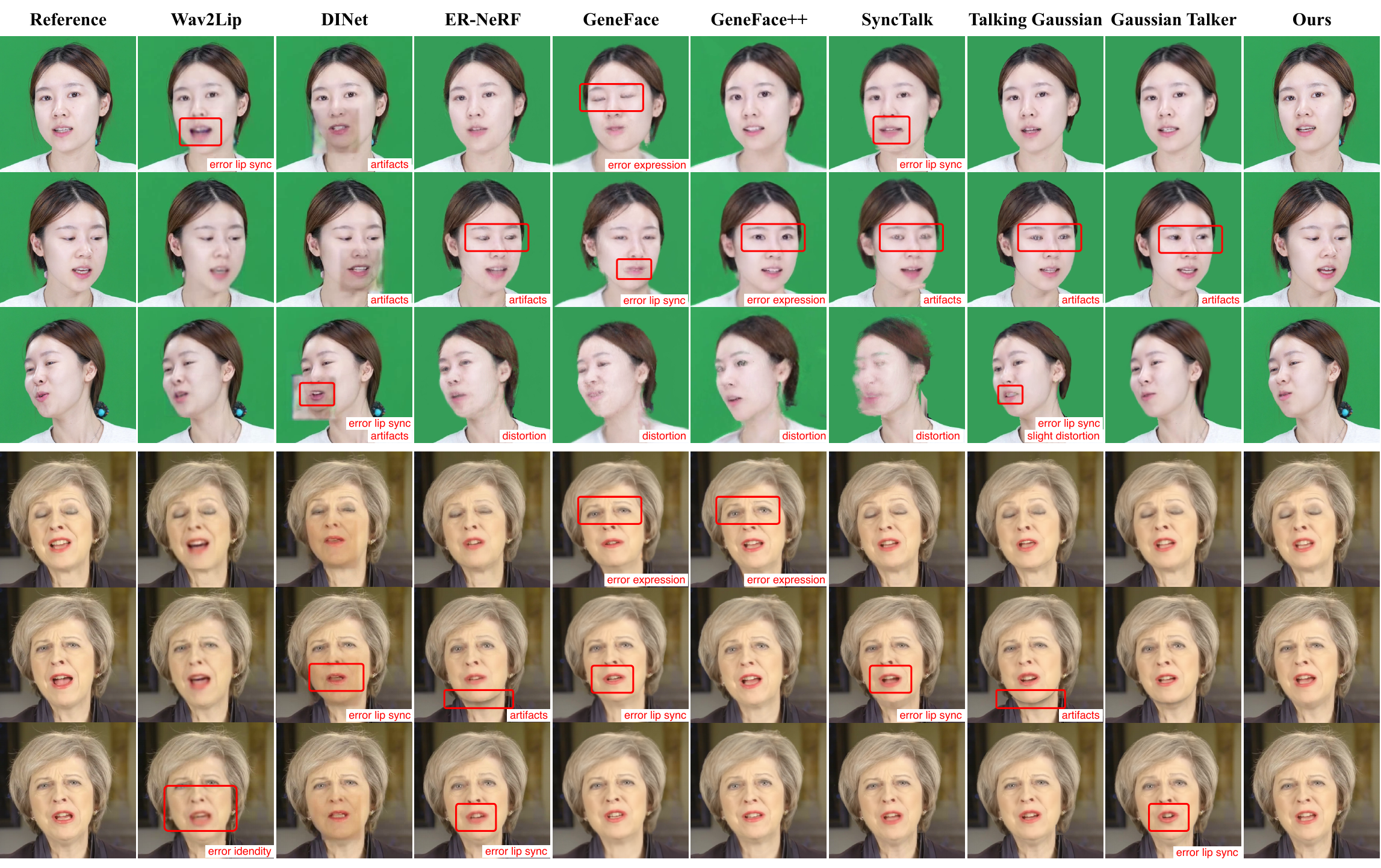}
\end{center}
\vspace{-1em}
   \caption{\textbf{Qualitative comparison of self-reenactment with previous methods.} Compared to previous methods, our method maintains excellent identity consistency and rendering quality across various viewing angles. Our method also provides more accurate lip-sync and expression transfer. Better zoom in for more details.}
\label{fig:main_compare}
\vspace{-1em}
\end{figure*}

\section{Experiments}
\footnotetext{Wav2Lip is jointly trained with SyncNet, using LSE-C as its optimization objectives. As a result, it obtains better scores than the ground truth.}

\subsection{Experimental Settings}
\noindent\textbf{Dataset.}
We train our Audio-Expression Priors on HDTF~\cite{zhang2021flow}, CN-CVS~\cite{chen2023cn} and self-collected 100-hour internet speech videos. We train our Expression-Visual Priors module on VFHQ~\cite{xie2022vfhq} and NeRSemble~\cite{kirschstein2023nersemble}. In NeRSemble, only front-facing frames are used for source image. For identity-specific test, we use commonly adopted datasets from previous works~\cite{guo2021ad, li2023efficient, ye2023geneface, peng2024synctalk} along with self-recorded videos to ensure a fair and comprehensive comparison. Specifically, the previous datasets include four video clips where the subjects speak while facing forward. In contrast, our two recorded videos feature larger head movements, better reflecting real-world application scenarios. The training videos include English, Chinese and French. All videos have a frame rate of 25 FPS and a duration of 4 to 6 minutes. Except for the video from AD-NeRF~\cite{guo2021ad}, which has a resolution of $450\times450$, all other videos have a resolution of $512\times512$, with the head-centered.

\noindent\textbf{Baselines.}
We compare GGTalker with previous methods, including GAN-based methods: Wav2Lip~\cite{prajwal2020lip}, DINet~\cite{zhang2023dinet} and IP-LAP~\cite{zhong2023identity}; NeRF-based methods: AD-NeRF~\cite{guo2021ad}, RAD-NeRF~\cite{tang2022real}, ER-NeRF~\cite{li2023efficient}, GeneFace~\cite{ye2023geneface}, Geneface++~\cite{ye2023geneface++} and SyncTalk~\cite{peng2024synctalk}; and 3DGS-based methods: TalkingGaussian~\cite{li2024talkinggaussian} and GaussianTalker~\cite{cho2024gaussiantalker}. 

\noindent\textbf{Implementation Details.} 
We use VHAP~\cite{qian2024versatile} to extract FLAME and camera parameters from both monocular and multi-view videos. We use the Adam optimizer~\cite{kingma2014adam} with a learning rate of $10^{-4}$ during the prior stages and $10^{-5}$ during the adaptation stage. Both prior stages are trained on 8 A100 GPUs, each taking about 2 days. Customized Adaptation is performed on a A100 GPU and takes around 20 minutes. For detailed fine-tuning steps and the time required for each step, please refer to the supplementary material.

\subsection{Quantitative Evaluation}
\noindent\textbf{Metrics.} 
In the self-reenactment setting, we measure image quality by calculating similarity metrics between the rendering results and source videos: \textbf{PSNR}, \textbf{LPIPS}~\cite{zhang2018unreasonable}, \textbf{SSIM}, and \textbf{FID}~\cite{heusel2017gans}. We measure the accuracy of facial movements with landMark distance (\textbf{LMD}) and action units error (\textbf{AUE})~\cite{baltruvsaitis2015cross}. To evaluate the synchronization between lip movements and audio, we also introduce Lip Synchronization Error Distance (\textbf{LSE-D}) and Confidence (\textbf{LSE-C})~\cite{prajwal2020lip}. Finally, we compare the \textbf{training time} and frames-per-second (\textbf{FPS}) as measures to evaluate the efficiency of each method. Since our method synthesizes only the head region, for a fair comparison, we measure image reconstruction metrics solely within the head region.


\noindent\textbf{Evaluation Results.} Tab.~\ref{tab:main_compare} shows the evaluation results of self-reenactment. Our method significantly outperforms previous approaches in facial quality metrics. This is because our rigged Gaussians have a well-defined 3D meaning, ensuring identity consistency and rendering quality from various angles. Thanks to the explicit control of pose and blinking through FLAME coefficients, we achieve better results in LMD and AUE metrics compared to prior methods. To more comprehensively evaluate lip-syncing capabilities, we compare our method with the latest SOTA approaches using OOD audio in Tab.~\ref{tab:OOD_compare}. We divide the OOD audio into two categories: cross-identity and cross-language. In the cross-identity setting, we use audio with the same language as the training video but with a different speaker. In the cross-language setting, the talking heads are driven by speech with both different languages and different voice characteristics. In both scenarios, GGTalker demonstrates state-of-the-art lip-syncing performance, overcoming the limited audio generalization issue seen in previous methods. Moreover, our method significantly outperforms previous approaches in training time and inference speed, greatly reducing the computational cost of creating talking heads and highlighting the vast application potential of GGTalker.

\subsection{Qualitative Evaluation}
\noindent\textbf{Evaluation Results.} 
To enable a more intuitive comparison, we present the results of qualitative experiments in Fig.~\ref{fig:main_compare}. 2D-based methods exhibit poor identity consistency and image quality. For example, Wav2Lip~\cite{prajwal2020lip} produces blurry lips, while DINet~\cite{zhang2023dinet} generates noticeable artifacts in the facial patch regions. Almost all 3D-based methods suffer from artifacts (especially around the eyes) or color distortion across the face during large head movements. This is because NeRF and 3DGS have high degrees of freedom and flexibility, making them prone to overfitting to the primary viewing direction without proper spatial constraints and initialization. In contrast, our method maintains excellent identity consistency and rendering quality across various viewing angles. Additionally, compared to ER-NeRF~\cite{li2023efficient} and TalkingGaussian~\cite{li2024talkinggaussian}, our approach avoids neck artifacts that often arise from hard stitching. Compared to GeneFace~\cite{ye2023geneface} and GeneFace++~\cite{ye2023geneface++}, we achieve more precise expression control through the explicit control of blinking with FLAME. Furthermore, by learning Audio-Expression priors from large-scale datasets and fine-tuning for the speaking style of a specific identity, we also achieve superior lip-sync accuracy. We recommend watching the supplementary video for better comparison.

\begin{table}[]
\setlength\tabcolsep{3pt}
\begin{center}
\resizebox{\linewidth}{!}{
        \begin{tabular}{lccc}
        \toprule
         Method& LPIPS ↓ & LMD ↓ & LSE-C ↑\\
        \midrule
         \textbf{ours}& \textbf{0.0281}& \textbf{2.328}& \textbf{5.769}\\
  w/o Audio-Expression Priors& 0.0306& 2.741&3.268\\
         w/o Audio-Expression Fine-tuning& 0.0385& 3.287& 4.780\\
  w/o Expression-Visual Priors& 0.0438& 3.826&4.682\\
 w/o Expression-Visual Fine-tuning& 0.0473& 2.925&4.524\\
         w/o Fine-tuning Gaussian color& 0.0336& 2.470& 5.431\\
        \bottomrule
        \end{tabular}}
\end{center}
\vspace{-1em}
\caption{\textbf{Quantitative results of ablations.}  We evaluate LPIPS and LMD for self-reenacment, LSE-C for cross-language tasks. }
\label{tab:ablation}
\vspace{-1em}
\end{table}

\begin{figure}
\begin{center}
   \includegraphics[width=1.\linewidth]{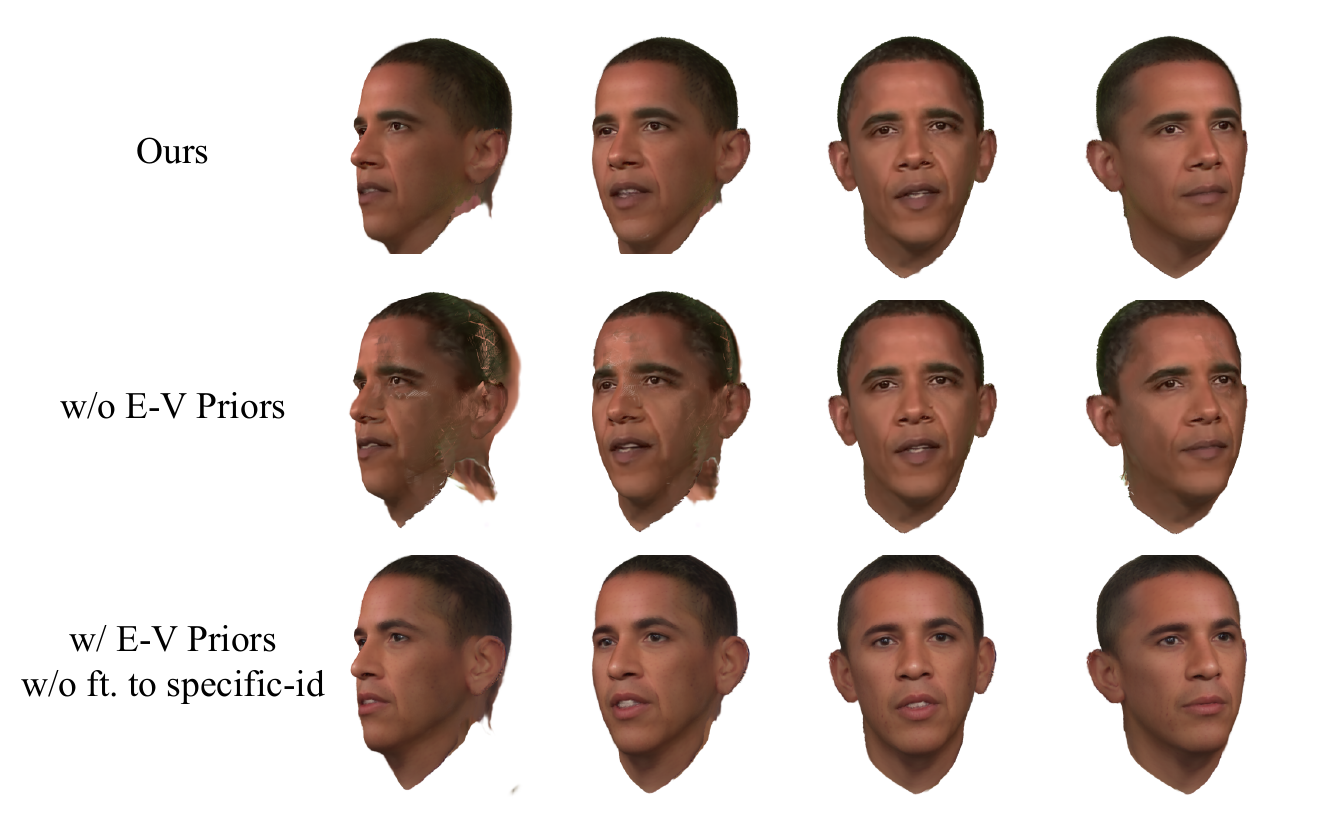}
\end{center}
\vspace{-1em}
   \caption{\textbf{Qualitative ablations on Expression-Visual.} We present a comparison of rendered results from novel views to demonstrate the contribution of our two-stage training. Removing either the prior or adaptation stage results in visual artifacts. }
\label{fig:ablation}
\vspace{-1em}
\end{figure}

\subsection{Ablations}
We conduct a series of ablation experiments to examine the effectiveness of each component of our method. To measure the image quality and generalizability, we calculate LPIPS and LMD for self-reenactment, and LSE-C for cross-language reenactment. 

Tab.~\ref{tab:ablation} shows the quantitative results of our ablations. The prior and fine-tuning stages of Audio-Expression introduce generalizability to multiple languages and identity-specific speaking styles, respectively. Removing the prior stage (i.e., training from scratch for the target subject) results in an inability to adapt to OOD audio, similar to other identity-specific 3D methods~\cite{li2023efficient, ye2023geneface, li2024talkinggaussian, cho2024gaussiantalker}. On the other hand, removing the fine-tuning stage leads to unnatural speaking styles, reflected in a slight drop in LSE-C. Fig.~\ref{fig:ablation} illustrates the contributions of the two Expression-Visual stages through novel-view rendering. Training from scratch lacks 3D consistency, causing holes and artifacts in novel views. Without fine-tuning, the head texture appears overly smooth, resulting in visually incorrect identity representation. Additionally, fine-tuning the Gaussians' color based on FLAME parameters leads to sharper textures. When removing this module, all metrics show a slight decline.

%% file: sec/5_conclusion.tex
\section{Conclusion}
In this paper, we propose GGTalker, a novel pipeline for high-fidelity, generalizable and rapid talking head synthesis. We demonstrate that introducing head priors learned from large-scale datasets into the adaptation of specific identities can significantly enhance the generalization of synthesized talking heads. We separately train Audio-Expression and Expression-Visual priors to learn a unified paradigm of lip movements and the general patterns of head textures. During Customized Adaptation, we adapt to each individual’s speaking style and texture details, introducing a color MLP and Body Inpainter to produce highly realistic results. Extensive experiments show that GGTalker achieves state-of-the-art visual quality and 3D consistency while maintaining precise lip-sync even for OOD audio. We believe that GGTalker's generalizability will further expand the real-world applications of talking head synthesis. 

%% file: sec/6_Acknowledgement.tex
\section{Acknowledgement}
We would like to thank Ms. Yujia Zhai and Ms. Chengbei Zou for providing some of the portraits and voices, respectively, for the showcase video!